\documentclass[conference]{IEEEtran}
\IEEEoverridecommandlockouts

\usepackage{amsmath,amssymb,amsfonts}
\usepackage{algorithmic}
\usepackage{graphicx}
\usepackage{textcomp}
\usepackage{xcolor}
\usepackage{tabularx}
\usepackage{graphicx}
\usepackage{color}
\usepackage{comment}
\usepackage{array}
\usepackage{hyperref}
\usepackage{booktabs}
\usepackage{multirow}

\begin{document}

%\title{French Clinical Document Analysis Using Machine Learning: ICD-10 Code Association\\
%\JFC{Titre : ICD-10 Code Association, French ...}
\title{Automatic ICD-10 Code Association: A Challenging Task on French Clinical Texts}

\author{\IEEEauthorblockN{Yakini Tchouka, Jean-François Couchot, David Laiymani}
\IEEEauthorblockA{
\textit{Université de Franche-Comté}\\
\textit{CNRS, Institut FEMTO-ST}\\
F-25000 Besançon, France\\
firstname.name@femto-st.fr}
\and

%\IEEEauthorblockN{Jean-François Couchot}
%\IEEEauthorblockA{
%\textit{Université de Franche-Comté}\\
%\textit{CNRS, Institut FEMTO-ST}\\
%F-25000 Besançon, France\\
%jfcouchot@femto-st.fr}
%\and

%\IEEEauthorblockN{David Laiymani}
%\IEEEauthorblockA{
%\textit{Université de Franche-Comté}\\
%\textit{CNRS, Institut FEMTO-ST}\\
% F-90000 Belfort, France\\
%david.laiymani@univ-fcomte.fr}

\IEEEauthorblockN{Philippe Selles, Azzedine Rahmani}
\IEEEauthorblockA{
\textit{Nord Franche-Comté Hospital}\\
F-90400 Trevenans, France\\
firstname.name@hnfc.fr}
\and

%\IEEEauthorblockN{Azzedine Rahmani}
%\IEEEauthorblockA{\textit{Nord Franche-Comté Hospital}\\
%F-90400 Trevenans, France\\
%azzedine.rahmani@hnfc.fr}

}
\maketitle
\begin{abstract}
Automatically associating ICD codes with electronic health data is a well-known NLP task in medical research.  NLP has evolved significantly in recent years with the emergence of pre-trained language models based on Transformers architecture, mainly in the English language. This paper adapts these models to automatically associate the ICD codes. Several neural network architectures have been experimented with to address the challenges of dealing with a large set of both input tokens and labels to be guessed. In this paper, we propose a model that combines the latest advances in NLP and multi-label classification for ICD-10 code association. Fair experiments on a Clinical dataset in the French language show that our approach increases the $F_1$-score metric by more than 55\% compared to state-of-the-art results.
\end{abstract}

\begin{IEEEkeywords}
natural language processing, icd-10, clinical document, unstructured data, multi-label classification, supervised learning, health, transformers
\end{IEEEkeywords}

\section{Introduction}
For a more accurate long term follow-up, patient's stay in a health center is usually reported in a digital documents which constitute the patient's medical record. Written by the patient's physicians, it is composed of operating reports, clinical notes, liaison letters etc. 
In a large number of countries each patient record is then classified according to the International Classification of Diseases (ICD). 
ICD is a medical classification system used for coding diseases and other health-related conditions. 
It is maintained by the World Health Organization (WHO) and is widely used globally as a standard for tracking health statistics, billing for medical services, and conducting research. 
In its $10^{th}$ edition \cite{world1992icd}, ICD is organized into chapters based on different body systems and disease categories. 
The chapters are further divided into subcategories that provide more specific information about the condition being coded. 
Each code consists of an alphanumeric string that includes a category code, a subcategory code, and a descriptor code (up to seven characters). 
The ICD-10 classification system is used by healthcare providers and organizations worldwide to standardize the coding of medical conditions and facilitate the sharing of health information across different systems and platforms. 
This classification is the common foundation for epidemiology, public health research, and healthcare management. 
In addition, the reimbursement of medical expenses by public or private insurance companies directly depends on the codes associated with these medical records. This makes it even more important to associate the right codes with each patient's record.
Finally, it should be noted that a more or less complex patient record can generate several ICD-10 codes

Typically, in a hospital, the responsibility for the ICD-10 classification falls on the medical coders. 
Staff performing this task is specially trained professionals who use medical documentation to assign the appropriate ICD-10 codes to medical records. 
Medical coders work closely with healthcare providers, nurses, and other staff to ensure that the medical records are accurately encoded into this classification.
In some hospitals the ICD-10 classification is performed by the physicians. However, regardless of how medical coding is managed, accuracy, and attention to detail are crucial to ensure that the data generated is reliable and useful for patient care and management. That is why automatically associating ICD codes to a medical record is a task that has been widely addressed in medical research in recent years \cite{choi2016doctor, baumel2018multi, vu2020label, dalloux2020supervised, huang2022plm}.

With the recent advances in Natural Language Processing (NLP) and since medical records are unstructured medical documents, 
it makes sense to apply these theoretical and technological advances in the context of ICD-10 classification. 
Clearly, the emergence of the Transformers architecture \cite{vaswani2017attention, devlin2018bert} has taken natural language processing to a new precision level. 
Several works have shown that the representations produced by these models are the most accurate and it is the most used architecture today in a large number of machine learning tasks (from text, to computer vision and time series) \cite{vaswani2017attention, khan2022transformers, lin2022survey}.
%\JFC{ on dit plusieurs travaux, mais il n'y a qu'une seule référence à la phrase précédente}

Nevertheless, ICD-10 automatic classification is a multi-label text classification task with tough challenges. For instance, the ICD-10 classification consists of about 140,000 codes (procedure codes and medical codes).
Unless one has a huge dataset, extremely important physical resources, and an extremely long period of time, it seems to be unrealistic to believe that one could associate to a patient record one of the 140,000 existing codes with a high degree of accuracy.
%\JFC{vérifier la phrase précédente que je viens de pondre. Une ref disant ceci serait la bienvenue} 

This large number of labels clearly stresses existing deep learning models to their limits. 
Another challenge is the size of the medical notes which far exceeds the usual limit of transformer architectures (typically $512$ tokens). 
Finally, working on non-English data is also challenging since the vast majority of open-source models available are trained on English corpus.

In this paper, we propose to address the three previous challenges for the ICD-10 classification of French medical records. 
We developed a deep learning model that combines the latest advances in Natural Language Processing.
This approach makes it possible to associate a non-negligible part of the existing ICD-10 codes on French-language patient records with an $F_1$-score outperforming with more than 55\% latest state of the art approach.
%It allows to classify a larand which is based on a number of relevant ICD-10 codes. We achieve the highest results to date in the French language with a $F_1$-score of $0.55$ for 1.564 codes and of $0.45$ for 6.610 codes.

This paper is organized as follows. 
Section~\ref{sec:art} starts with recalling state of the art of associating ICD codes to medical records. 
Section~\ref{sec:dataset} presents the dataset used on the one hand to validate our approach and on the other hand to fairly compare the $F_1$-scores obtained by our approach with those obtained with already existing approaches.
The architecture of our ICD code association model is presented in Section \ref{sec:arch}.
Results are presented and analyzed in Section \ref{sec:eval}.
Concluding remarks and future work are finally given in Section~\ref{sec:concl}.

\section{Related Work}\label{sec:art}
\subsection{Natural Language Processing}
NLP has significantly evolved in recent years with the joint appearance of the Transformers model \cite{vaswani2017attention} and their generalization ability to transfer learning. ELMo \cite{peters-etal-2018-deep} and BERT \cite{devlin2018bert} have shown this effectiveness which provides more accurate contextualized representations. Several pre-trained models then appeared such as BERT, RoBERTa \cite{liu2019roberta} \ldots. These models are pre-trained on a large amount of general domain English text to capture the ability to model text data, and then refined on common classification tasks. In French two main models have been proposed i.e FlauBERT \cite{le2019flaubert}, CamemBERT \cite{martin2019camembert}. Note that some multi-lingual models also exist such as XLM-R \cite{xlm-r}. Some models are also trained on domain-specific text corpus. For example, ClinicalBERT\cite{alsentzer2019publicly} and BioBERT\cite{lee2020biobert} have been trained on medical data to address medical domain tasks. Unfortunately, there is no such model in the French language, leading to a gap between the usage of machine learning approaches on French documents compared to the same approach in English ones. In general, Transformers models have a limited input size ($512$ tokens in pratice).
In the case of clinical documents this limit can become very penalizing since a typical patient document is generally much larger than $512$ words or tokens. In \cite{pappagari2019hierarchical} the authors proposed some hierarchical methods to tackle this problem. They divided the document into several segments that can be processed by a Transformers. Then the encoding of the segments is aggregated into the next layer (Linear, recurrent neural networks or other layer of Transformers). Recently, the sparse-attention system i.e. the \textit{LongFormer} model has been proposed in \cite{beltagy2020longformer}. It is composed of a local attention (attention between a window of neighbour tokens) and a global attention that reduces the computational complexity of the model. They can therefore be deployed to process up to 4096 tokens.

\subsection{ICD Code Association}
The automatic association of ICD codes is one of the most addressed challenges in medical research. With the emergence of neural networks and the evolution of natural language processing, several authors have tried to tackle this task. \cite{choi2016doctor} and \cite{baumel2018multi} used recurrent neural networks (RNNs) to encode Electronic Health Records (EHR) and predict diagnostic outcomes. On the other hand, \cite{shi2017towards} and \cite{mullenbach2018explainable} have used the attention mecanism with RNNs and CNNs to implement more accurate models.

%%citep
The work of \cite{xie2018neural} and \cite{tsai2019leveraging} present various ways to consider the hierarchical structure of codes. \cite{xie2018neural} used a sequence tree LSTM to capture the hierarchical relationship between codes and the semantics of each code. \cite{cao2020hypercore} proposed to train the integration of ICD codes in a hyperbolic space to model the code hierarchy. They used a graph neural network to capture code co-occurrences. LAAT \cite{vu2020label} integrated a bidirectional LSTM with an attention mechanism that incorporates labels.

EffectiveCAN \cite{liu2021effective} used a squeeze-and-excitation network and residual connections as well as extraction of representations from all layers of the encoder for label attention. The authors also introduced focal loss to address the problem of long-tail prediction with $58.9$\% of $F_1$-score on MIMIC 3 \cite{johnson2016mimic}. ISD \cite{zhou2021automatic} used shared representation extraction between high frequency layers and low frequency layers and a self-distillation learning mechanism to mitigate the distribution of long-tailed codes.

Recently \cite{huang2022plm} proposed the PLM-ICD system that focuses on document encoding with multi-label classification. They used an encoding model based on the Transformers architecture adapted to the medical corpus. Associating ICD-10 codes is finding the codes corresponding to medical documents in a large set of codes. For instance MIMIC 3 \cite{johnson2016mimic} contains more than 8,000 codes, and handling such large set of labels in classification is a challenging problem in machine learning. To overcome this problem, the authors used the Label-Aware Attention (LAAT) mechanism proposed in \cite{vu2020label} which integrates labels in the encoding of documents. Finally, to solve the problem of long sequences they used the hierarchical method. PLM-ICD is the current state-of-the-art model that achieved $59.8$\% of $F_1$-score on MIMIC 3 \cite{johnson2016mimic} and $50.4$\% on MIMIC 2 \cite{saeed2011multiparameter}.

In French, \cite{dalloux2020supervised} proposed Convolutional Neural Networks (CNN) models with multi-label classification to automatically associate ICD-10 code. The authors used FastText \cite{fasttext} vectors with the skip-Gram algorithm for the encoding of documents. They first considered all the final labels of the dataset, then grouped them into families to reduce the number of classes. This model is trained on a private dataset of 28,000 clinical documents and reached $39$\% of $F_1$-score with 6,116 codes and $52$\% with 1,549 codes.

\section{Dataset}\label{sec:dataset}
This work is in collaboration with The Hopital Nord Franche-Comté (HNFC), a French public health center that provided us with patient stays. For privacy reasons, all our experiments were conducted on site and no data was taken out of the hospital.
 
\begin{figure}[htbp]
    \centering
    \includegraphics[width=\linewidth]{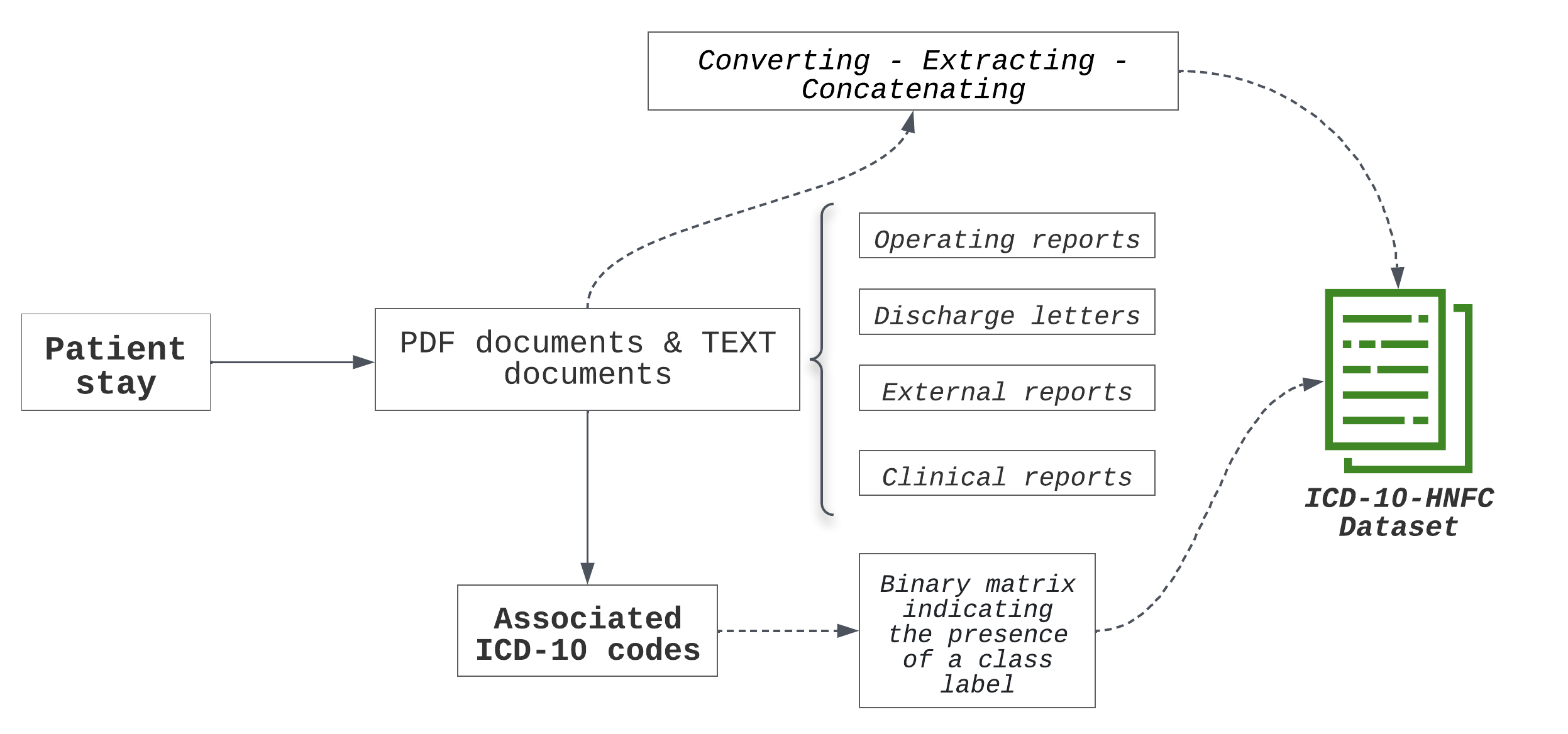}
    \caption{ICD-10-HNFC Dataset Construction}
    \label{dataset}
\end{figure}

A patient's stay is a set of successive visits in possibly different departments of the hospital. Each department produces a clinical document that describes the patient's stay in that department. These clinical documents are used by the medical coding specialists to associate the corresponding ICD-10 codes. We finally obtain a set of unstructured textual documents corresponding to the global stay of the patient to which a set of codes is associated. As clinical documents, we have for example operating reports, discharge letters, external reports or clinical notes. The obtained dataset, further denoted as ICD-10-HNFC dataset is therefore a database of groups of medical documents with associated codes. This system is well illustrated in Fig.~\ref{dataset}.

ICD-10-HNFC dataset is built for supervised deep learning. In supervised learning, to have an accurate model, there are several factors to consider. Is there enough training data? Is the number of classes consistent with the volume of data available? Is the frequency of classes in the dataset balanced? It is always difficult to find the perfect dataset that answers all these questions. In this paper, we worked not only on the main dataset, which consists in associating the raw ICD codes it contains but also on the sub-datasets such as associating the most frequent codes or code families instead of the raw codes.

\subsection*{Class Reduction}
As mentioned, the ICD is a classification system that is composed of thousands of codes. Given the large number of labels present in our basic dataset (shown in Table \ref{tab:label}), it is difficult to approach this classification task by considering all the classes present.  By doing so, the results of the constructed model will be far from perfect. The most precise models to date in English for ICD-10 code association is PLM-ICD which reached $59.8$\% on MIMIC 3 with 8,922 labels \cite{johnson2016mimic} and $50.4$\% on MIMIC 2 with 5,031 labels \cite{saeed2011multiparameter}. This proves the difficulty of this task. The first sub-dataset consists in reducing the codes to the first $3$ characters seen as a family. Therefore, instead of considering the raw codes, we will group them into families. This reduces in a consequent way the number of classes to be treated by the model. This dataset is presented in Table \ref{tab:label}. We can see via the description "line Code with less than 10 examples" in Table \ref{tab:label} that the reduction of the classes not only allows to have a more reasonable number of classes but also increase the frequency of the codes in the dataset.
\subsection*{Code Frequency}

Associating ICD-10 codes is a very frequent task in health centers. As a result, some codes occur more frequently than others. Finding the most frequent codes automatically can only be useful. Our second sub-dataset consists in building models based on the number of codes ($K$) that we consider more relevant. We evaluate the relevance based on the frequency of the code in the dataset. Thus, a model built on such a dataset will be able to associate the integrated codes with a better classification performance.

\subsection*{Additional Label}

With the code frequency strategy ($K$), the dataset is therefore composed of entries whose association belongs only to the $K$ most relevant codes. To keep the coherence of our dataset an additional label is introduced to represent the codes which are not considered as relevant (least frequent codes). So instead of having $K$ classes, the model will have $K+1$ ones. The additional class mean concretely in the association that there are one or more additional codes to associate. 

\begin{table}[htbp]
    \caption{Descriptive statistics of ICD-10-HNFC dataset}
    \centering
    \begin{tabular}{|c|c|c|}
    \hline
     & \multirow{2}{*}{Dataset} & Dataset\\
     &  & with class reduction\\
    \hline
    Documents & 56014 & - \\
    \hline
    Tokens  & 41868993 & - \\
    \hline
    Average sequence length  & 747 & -\\
    \hline
    Total ICD codes  & 416125 & 415830\\
    \hline
    Unique ICD codes & \textbf{6160} & \textbf{1564}\\
    \hline
    Codes with less than 10 examples & 3722 & 523\\
    \hline
    Codes with 100 examples or more & 641 & 471\\
    \hline
    \end{tabular}
    \label{tab:label}
\end{table}

%\DL{DL: il faut préciser d'où vient le dataset et parler du partenariat avec l'HNFC (sûrement reprendre ce que l'on a dit pour les autres publis) et parler de l'anonymization}

%\section{METHODOLOGY}\label{sec:meth}

%The ICD-10 code association task is equivalent to multi-label classification by considering the codes as the classes of our model. The data being textual, it is therefore the same as doing textual classification. As mentioned before, the best systems are those based on deep learning and using the latest textual representation systems (Transformers\cite{vaswani2017attention}). In this work we use supervised learning to set up our association model. 

\section{Model Architecture}\label{sec:arch}

This section presents the different components of the model architecture we have developed and justifies the choices made to design it. As previously exposed,  as we deal with the French language our choice was to fine-tune pre-trained transformer-based French models i.e. CamemBERT \cite{martin2019camembert} and FlauBERT \cite{le2019flaubert} for the implementation of the model architecture. 

%Unfortunately, the average number of tokens of our inputs exceeds the maximum limit of these models (i.e. $512$ tokens). It is thus required to add a long sequence processing system to get the global representation of the inputs. We 

%Recently, \citep{dai2022revisiting} summarized the available methods for processing long sequences via transformers.

\subsection{Global Document Representation}

As mentioned, Transformers main constraint is the limitation of the number of tokens present in an input sequence. Since the average size of the clinical notes of ICD-10-HNFC dataset exceeds this limit ($747$ versus $512$ as shown in Table \ref{tab:label}), basic Transformers cannot be used. Recently, \cite{dai2022revisiting} summarized the available methods for processing long sequences via Transformers. They can be summarized as hierarchical Transformers and sparse-attention Transformers in which we can find the \textit{Longformer} model of \cite{beltagy2020longformer} early mentioned. \textit{Longformer} can process up to $4096$ tokens per sequence allows to meet this limit. Unfortunately, there is no French pre-trained \textit{Longformer} model to date. Therefore, in this paper, we will use the hierarchical method to tackle this problem.

Hierarchical Transformers\cite{pappagari2019hierarchical, dai2022revisiting} are built on top of Transformers architecture. A document $D$ , is first divided into segments $[t_0, t_1, \dots , t_{|D|}]$, each of which must have less than $512$ tokens (the limit of Transformers). These segments are encoded independently using a typically pre-trained Transformers. We then obtain a list of segment representations which must be aggregated to obtain the whole document $D$ representation. There are several ways to do this aggregation. The aggregator can be an average of the representations of all the segments of the document (mean pooling) or the maximum of the representations in each dimension of the segments (max pooling) or stacking the segment representations into a single sequence. The aggregated sequence thus serves as an input to the next layer.

\subsubsection{Classification of a Large Number of Labels}

To overcome the problem of a large set of labels since ICD-10-HNFC contains more than 6,000 codes, we used the Label-Aware Attention (LAAT) system as in \cite{huang2022plm}. LAAT consists in integrating the labels into the document representation. Label-Aware Attention captures important text fragments related to certain labels. Let $H$ be the stacking representation of an input sequence. First, a label-wise attention weight matrix $Z$ is computed as follows: 

\begin{equation*}
    Z = tanh(VH)
\end{equation*}
\begin{equation*}
    A = softmax(WZ)
\end{equation*}
where $V$ and $W$ are linear transforms. The $i^{th}$ row of $A$ represents the weights of the $i^{th}$ label. The softmax function is performed for each label to form a distribution over all tokens. Then, the matrix $A$ is used to perform a weighted-sum of $H$ to compute the label-specific document representation:
\begin{equation*}
    D = HA^T
\end{equation*}

The $i^{th}$ row of $D$ represents the document representations
for the $i^{th}$ label. Finally, $D$ is used to make predictions by computing the inner product between each row of $D$ and the related label vector. 

\begin{figure*}[htpb]
    \centering
    \includegraphics[width=0.9\linewidth]{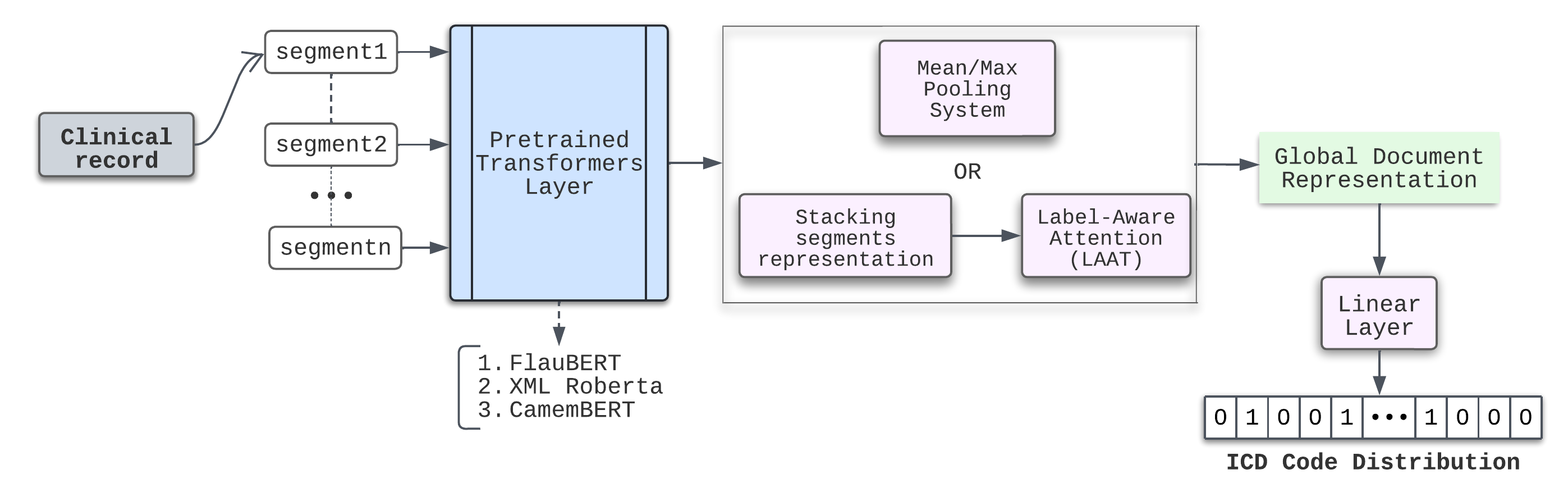}
    \caption{Global Architecture}
    \label{architecture}
\end{figure*}

In this paper, several architectures were experimented such as the model without long sequence processing, the model with long sequence processing (max/mean pooling), and the model with LAAT. The global architecture is illustrated in Fig.~\ref{architecture}.

\section{Experiments and Analysis}\label{sec:eval}
In this section, we present the results of the experiments conducted with the previously detailed architectures and dataset. We compare the results of recent works (PLM-ICD\cite{huang2022plm}, CNN\cite{dalloux2020supervised}) on the association of ICD-10 codes with 

\begin{table}[htbp]
    \caption{ICD-10 association results of the different architectures on the validation ICD-10-HNFC dataset}
    \centering
\setlength{\tabcolsep}{3pt}
    \begin{tabular}{|c|c|c|c|c|}
        \hline
    Models & Labels & Precision & Recall & $F_1$-score\\
    \hline
    FlauBERT (512 tokens) & \multirow{5}{*}{1564} & 0.48 & 0.31 & 0.38\\
    Hierarchical Mean FlauBERT & & 0.54 & 0.39 & 0.45\\
    Hierarchical Max FlauBERT & & 0.53 & 0.40 & 0.46\\
    FlauBERT + LAAT & & \textbf{0.57} & 0.51 & 0.54\\
    CamemBERT + LAAT & & 0.56 & \textbf{0.53} & \textbf{0.55} \\
    \hline
    
    FlauBERT + LAAT &\multirow{2}{*}{6160} & 0.41 & 0.43 & 0.42 \\
    CamemBERT + LAAT &  & 0.52 & 0.4 & \textbf{0.45} \\
    \hline
    \end{tabular}
    
    \label{res:global}
\end{table}

the most precise model of this paper. To evaluate our model we use the most used performance measures in classification: Precision, Recall, $F_1$-score. The micro average system is used to obtain the aggregation of the performances.

\subsection{Paper Models}
First, we conducted the experiments on the ICD-10-HNFC dataset with class reduction ($1564$ labels) as detailed in Table \ref{dataset}. This experimentation is performed with all the architectures developed in this paper. They are listed in Table \ref{res:global}. Then we trained another model on the global ICD-10-HNFC dataset ($6101$ labels) with the architectures that obtained the highest $F_1$-score in the previous experiment. The results are shown in the Table \ref{res:global}. The results confirm the effects of the different components that constitute our architectures. In summary, the LAAT approach outperforms the hierarchical methods witch are better than the base truncated model.

\subsection{$K$-based Models}
As detailed in Section \ref{sec:dataset}, different models have been trained based on a number ($K$) of labels (i.e. the most frequent codes). We present here the evaluation of these models with $K$ in $[10, 50, 100, 200]$. As shown in Table \ref{tab:kres}, models are less and less accurate when we increase the number of labels (classes). This is simply due to the aggregation of performances.
The more different codes there are, the fewer instances of each code there are in the dataset, and the less easy the contextualization is.
%The more labels there are, the more the model will make erroneous classifications. 
\begin{table}[htbp]
    \caption{Results of the ICD-10 association of K-based models}
    \centering
    \begin{tabular}{|c|c|c|c|c|}
    \hline
    $K$ & Precision & Recall & $F_1$-score\\
    \hline
    10 & \textbf{84} & \textbf{80.5} & \textbf{82.1} \\
    50  & 78.2 & 65.1 & 71\\
    100 &  77.2 & 58.4 & 66.5\\
    200 &  71.9 & 52.6 & 60.8\\
    \hline
    \end{tabular}
    
    \label{tab:kres}
\end{table}
\subsection{Comparison with other Works}

The Table \ref{tab:comp} shows the model with the highest $F_1$-score of this paper with the results of previous work on ICD-10 code association. It is difficult to compare the results, since these works do not use the same evaluation dataset and English works can benefit from specialized models such as ClinicalBERT \cite{alsentzer2019publicly}. For French baseline, we implemented and trained the model proposed in \cite{dalloux2020supervised} on ICD-10-HNFC dataset. The result is shown in parallel with our proposal. Our model clearly outperforms the classification method used in \cite{dalloux2020supervised}. On the same validation dataset, with class reduction (1564 labels) the $F_1$-score goes from 0.35 obtained with the model proposed in \cite{dalloux2020supervised} to 0.55 with our proposal, i.e. an improvement of 57\%. With the raw codes (6161 labels), the $F_1$-score goes from 0.27 to 0.45, i.e. an improvement of 66.6\%. The difference in scores with the results of PLM-ICD can be explained by the use of a context specific (medical) Transformers which has a vocabulary more adapted to the content of the documents.

\begin{table}[htbp]
    \caption{Results comparison with the previous work on ICD-10 Association. The state of the art works with their results are in \textit{italic}. The experiments done in this paper with ICD-10-HNFC dataset are presented in the other part. The highest scores in each part in relation to the number of labels are marked in \textbf{bold}}
    \centering

    \begin{tabular}{|c|c|c|c|c|}
    \hline
    Models & Language & Dataset & Labels & $F_1$-score\\
    \hline
    \textit{\multirow{2}{*}{PLM-ICD\cite{huang2022plm}}} &
    \textit{\multirow{2}{*}{English}} &
    \textit{MIMIC 2\cite{saeed2011multiparameter}} & \textit{5,031} & \textit{0.5}\\
    & & \textit{MIMIC 3\cite{johnson2016mimic}} & \textit{8,922} & \textit{\textbf{0.59}}\\
    \hline
    
    \textit{\multirow{2}{*}{\cite{dalloux2020supervised}}} &
    \textit{\multirow{2}{*}{French}} &
    \textit{\multirow{2}{*}{\cite{dalloux2020supervised}}} & \textit{6,116} & \textit{0.39}\\
    & & & \textit{1,549} & \textit{0.52} \\
    \hline
    \multirow{2}{*}{\textbf{PROPOSAL}} &
    \multirow{4}{*}{French} &
    \multirow{4}{*}{ICD-10-HNFC} & 6,161 & \textbf{0.45}\\
    & & & 1,564 & \textbf{0.55} \\
    \cline{1-1}
    \cline{4-5}
    \multirow{2}{*}{\cite{dalloux2020supervised}} & \multirow{2}{*}{} & & 6,161 & 0.27\\
    & & & 1,564 & 0.35 \\
    \hline

    \end{tabular}
    
    \label{tab:comp}
\end{table}

\section{Conclusion}\label{sec:concl}
In this paper, we address the challenges of automatically associating ICD-10 codes to French clinical unstructured data. We have experimented several Transformers architectures to address the challenges of large input tokens and large numbers of labels. We therefore propose an ICD-10 association model that uses the latest advances in natural language processing and achieves the highest results in the French language to date. Our future work will focus on the use of Large Language Models and few-shots learning techniques to the ICD-10 classification.

\section*{Acknowledgment} 
This work is (partially) supported by the EIPHI Graduate School (contract ANR-17-EURE-0002).

\bibliographystyle{plain}
\bibliography{main}
\end{document}